# Comparative Analysis of Tokenization Algorithms for Low-Resource Language Dzongkha

Tandin Wangchuk[1][Orcid] and Tad Gonsalves[2][Orcid]

[1] Department of Information and Communication Science, Faculty of Science and Technology, Sophia University, Chiyoda-ku, Tokyo, 102-8554, Japan
`tandin@eagle.sophia.ac.jp`
[2] Department of Information and Communication Science, Faculty of Science and Technology, Sophia University, Chiyoda-ku, Tokyo, 102-8554, Japan
`t-gonsal@sophia.ac.jp`

**Abstract.** Large Language Models (LLMs) are gaining popularity and improving rapidly. Tokenizers are crucial components of natural language processing, especially for LLMs. Tokenizers break down input text into tokens that models can easily process while ensuring the text is accurately represented, capturing its meaning and structure. Effective tokenizers enhance the capabilities of LLMs by improving a model's understanding of context and semantics, ultimately leading to better performance in various downstream tasks, such as translation, classification, sentiment analysis, and text generation. Most pre-trained tokenizers are suitable for high-resource languages like English but perform poorly for low-resource languages. Dzongkha, Bhutan's national language spoken by around seven hundred thousand people, is a low-resource language, and its linguistic complexity poses unique NLP challenges. Despite some progress, significant research in Dzongkha NLP is lacking, particularly in tokenization. This study evaluates the training and performance of three common tokenization algorithms in comparison to other popular methods. Specifically, Byte-Pair Encoding (BPE), WordPiece, and SentencePiece (Unigram) were evaluated for their suitability for Dzongkha. Performance was assessed using metrics like Subword Fertility, Proportion of Continued Words, Normalized Sequence Length, and execution time. The results show that while all three algorithms demonstrate potential, SentencePiece is the most effective for Dzongkha tokenization, paving the way for further NLP advancements. This underscores the need for tailored approaches for low-resource languages and ongoing research. In this study, we presented three tokenization algorithms for Dzongkha, paving the way for building Dzongkha Large Language Models.

## 1 Introduction

Large Language Models (LLMs) are becoming increasingly popular and evolving fast. They have been helpful in all walks of life. In today's LLMs, tokenizers play important roles in natural language processing (NLP). Tokenizers segment input text into tokens that can be efficiently processed by models without degrading the accurate representation of the text, thus maintaining its meaning and structure. The effectiveness of input text processing relies on a well-designed tokenizer. A study by Ali et al. [1] illustrated that the choice of tokenizers can significantly impact downstream tasks for LLMs. Well-crafted tokenizers enhance the capabilities of LLMs by improving the model's understanding of semantics and context, leading to better performance in various downstream tasks, such as translation, classification, sentiment analysis, and text generation. Most pre-trained tokenizers perform well for high-resource languages like English but struggle with low-resource languages. Several studies [2], [3], [4] indicate that the tokenization process affects how effectively LLMs can perform in languages other than English.

Dzongkha, Bhutan's national language, spoken by approximately seven hundred thousand people, is regarded as one of the essential cultural identities for the country's survival. With changing times, the government recognized the need to modernize, leading to the initiation of several language computerization projects. Despite these initiatives, Dzongkha remains classified as one of the low-resource languages. Its linguistic complexity compounds unique NLP challenges. Although some progress [5], [6], [7], [8] has been observed in recent years, significant research in Dzongkha NLP, particularly in tokenization, is still lacking. In addition to the scarcity of training data, the absence of an efficient Dzongkha tokenizer hinders downstream classification tasks, such as the detection of phishing scams perpetrated recently through multilingual translation tools [9].

This study aimed to evaluate the initial vocabulary construction time on a Dzongkha corpus and the performance of three candidate tokenizing algorithms in comparison to other popular baseline algorithms. The performance was investigated based on Normalized Sequence Length, Subword Fertility, Proportion of Continuous Words, and execution time metrics on the given dataset. These metrics provide insights into the efficiency and effectiveness of different tokenizing algorithms which can inform future research and development in NLP for low-resource languages like Dzongkha. The rest of the paper is structured as follows: an Overview of Tokenization Algorithms, Evaluation Metrics, Experimental Setup, Results and Analysis, Discussion, and Conclusion.

## 2   Overview of Tokenization Algorithms

We chose three common subword tokenization frameworks for our study since they are the most popular and implemented in major LLMs. These are Byte-Pair Encoding, WordPiece and SentencePiece. For SentencePiece, we chose the Unigram implementation.

## 2.1 Byte-Pair Encoding (BPE)

Gage [10] initially proposed the Byte-Pair Encoding algorithm for data compression. Later this algorithm was adapted for character sequence tokenization and gained widespread use due to the work by Sennrich et al [11]. BPE is a compression technique adapted for word segmentation that iteratively merges a frequent pair of characters or sequences and replaces them with a new character or sequence. Ultimately a new vocabulary is created. New unseen sequences of characters or words are then tokenised based on this vocabulary. Tokenizers for models such as GPT2 to GPT4 [12], Llama 3 [12], RoBERTa [13], BART [14] and DeBERTa [15] are based on Byte-Pair Encoding. Subword tokenization has significantly improved the efficiency and accuracy of tokenizers in handling various NLP tasks for modern LLMs.

## 2.2 WordPiece

WordPiece is very similar to BPE and was first defined in the context of Japanese and Korean Voice Search [16]. BERT employed the WordPiece tokenization which constitutes the subword tokenization. It is also used in DistilBERT and Electra. In WordPeice, vocabulary is first initialized including every character present in the training corpus and subsequently learns a specified number of merge rules. Unlike BPE, WordPiece does not select the most frequently occurring character/sequence-pair but one that optimally enhances the likelihood of the training corpus when added to the vocabulary. The use of the WordPiece tokenizer has improved the performance of NLP especially in handling non-space-separated languages like Japanese and Korean.

## 2.3 SentencePiece (Unigram).

SentencePiece is an unsupervised model for text tokenization and detokenization, primarily designed for Neural Network-based text generation with a predetermined vocabulary size before training. According to the paper [17] and the GitHub page maintained for the paper, both Byte-Pair Encoding and Unigram language models are supported with the added feature of facilitating direct training on a raw corpus. SentencePiece is a language-independent, fast, and lightweight tokenization method that allows for efficient processing of raw text data.

## 3 Evaluation Metrics:

To investigate the performance of our candidate tokenizers, we selected four metrics: normalized sequence length, subword fertility, proportion of continued words and execution time. The Normalized Sequence Length (NSL) is the compression ratio of a tokenizer relative to a baseline tokenizer [18][19]. It measures the ratio of the length of the tokenized sequence generated by candidate tokenizers to the length of the sequence produced by the baseline tokenizer. For example, if the NSL value is 0.8, this means that on average, the candidate tokenizer uses 20% fewer tokens compared to the baseline tokenizer [19]. The subword fertility of a tokenizer is defined as a measure of the average number of tokens used to represent a word [3]. The ideal subword fertility value is 1 which indicates that the tokenizer's vocabulary contains every word in the input text[3], [18]. Proportion of Continued words measures the number of words fragmented into two or more tokens and the ideal value is 0.0[3]. The execution time is a measure of the time taken to tokenize a sequence of input text. The pre-training time of each candidate tokenizer is the time taken to train tokenizers on a given corpus. These metrics provide an extensive understanding of the efficacy and efficiency of various tokenizers which can inform future NLP research and development for low-resource language like Dzongkha.

## 4     Experimental Setup:

The dataset used for the evaluation of tokenizers was provided by DDC containing 180000 manually segregated Dzongkha words which were used for the master's thesis [8]. For this study, the dataset is pre-processed to extract only words and exclude markers such as "beg", "end", "mid", "#", "*", and "NUM". During the preprocessing, the count of words is maintained. The extracted words are concatenated to form a string to feed into the Tokenizers. This preprocessing step ensures that the dataset is clean and suitable for uniformly evaluating the performance of the various tokenization algorithms. Ensuring a consistent format can lead to more accurate and reliable results.

Before the evaluation, three tokenisation algorithms, namely WordPiece, Byte-Pair Encoding (BPE) and SentencePeice using Unigram, are pretrained with the corpus developed by DDC [20]. The corpus contains 27M Unicode characters. A uniform vocabulary size of 10000 and 30000 each is used to train the models from scratch. The trained models and vocabulary are saved locally for performance evaluations downstream. WordPiece and BPE tokenizers are realised using the Tokenizer library provided by HuggingFace. The Python Wrapper is used for the implementation of the SentencePiece tokenizer. The training time for each algorithm on the corpus is noted. The setup allows for a comparison of the performance of the different tokenization algorithms in an even and exhausted manner, with the models being trained and tested under the same conditions.

Using the trained models and vocabularies, the pre-processed dataset is used for the determination of Normalized Sequence Length, Sub-word Fertility and Proportion of Continued Words. The execution time for tokenisation by each algorithm is also recorded. These metrics are used for the evaluation of the algorithms. By analysing these performance metrics, each tokenization algorithm is comprehensively assessed to find out the most suitable algorithm for Dzongkha tokenization.

## 5     Results and Analysis:

WordPiece, BPE and SentencePiece are pretrained with the corpus of 27M characters developed by DDC with a uniform vocabulary size of 10000 and 30000. The training time of each algorithm is shown in Table 1. WordPiece and BPE take relatively faster time to train the corpus with 10000 or 30000 vocabulary sizes. Whereas SentencePiece is much slower than WordPiece and BPE in training either 10000 or 30000 vocab size for the corpus. Understanding the training times contributes to the evaluation of these algorithms for large corpus, especially for pretraining a Dzongkha LLM.

Table 1: Pretraining time for each candidate tokenizing algorithms

| Tokenization Algorithms | Avg. Time (Vocab Size: 10000) per loop | Avg. Time (Vocab Size: 30000) per loop | Corpus |
|---|---|---|---|
| WordPiece | 2.01 s ± 17.6 ms | 2.41 s ± 32.5 ms | 27 million Unicode characters |
| SentencePiece (Unigram) | 2 min 7s ± 3.57 s | 1min 48s ± 4.56 s | |
| Byte-Pair Encoding | 43 s ± 2.62 s | 46.5 s ± 1.98 s | |

Next, the tokenizers are evaluated for Normalized Sequence Length (NSL). NSL values provide a comparison of encoded sequences between our candidate and baseline tokenizers namely GPT2 tokenizer provided by HuggingFace, Llama 30B tokenizer and TikToken with 0200k_base encoding used for GPT-4o. NSL for each candidate tokenizer is calculated with respect to these three baseline tokenizers. Fig 1 shows the heatmap for the matrix of NSL values for each candidate algorithm evaluated with respect to three baseline tokenizers. SentencePiece with GPT2, Llama3 and TikToken has the lowest NSL values of 0.0594, 0.1162 and 0.1105 respectively. This indicates that the SentencePiece uses a smaller number of tokens than other candidate tokenizers. Analysis of these NSL values helps us to determine that SentencePiece is more efficient in terms of tokenization compared to other candidate tokenizers, making it a potentially better option for Dzongkha.

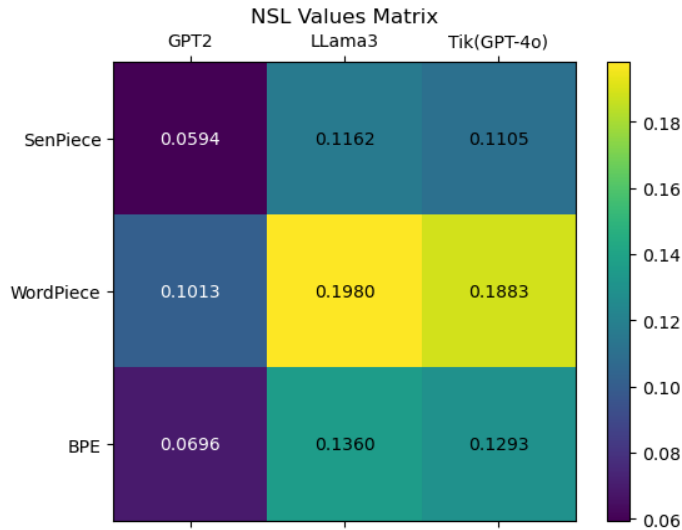

Fig. 1 Heatmap showing the matrix of NSL Values of various tokenizers

The third metric used for evaluation is Sub-word Fertility. Figure 2 displays the sub-word fertility of each tokenising algorithm. The ideal sub-word fertility value is 1.0 which indicates that, on average, each word is represented by a single token. Lower subword fertility generally signifies a more effective tokenizer, as it means that the input text can be converted into a shorter sequence of tokens. Conversely, a subword fertility significantly greater than 1.0, implies that the tokenizer is over-segmenting words into smaller units than necessary. Only SentencePiece has a value lower than 1.0. This analysis helps to identify the most efficient tokeniser in terms of word representation using tokens, with SentencePiece showing promising results by having a sub-word fertility value lower than 1.0.

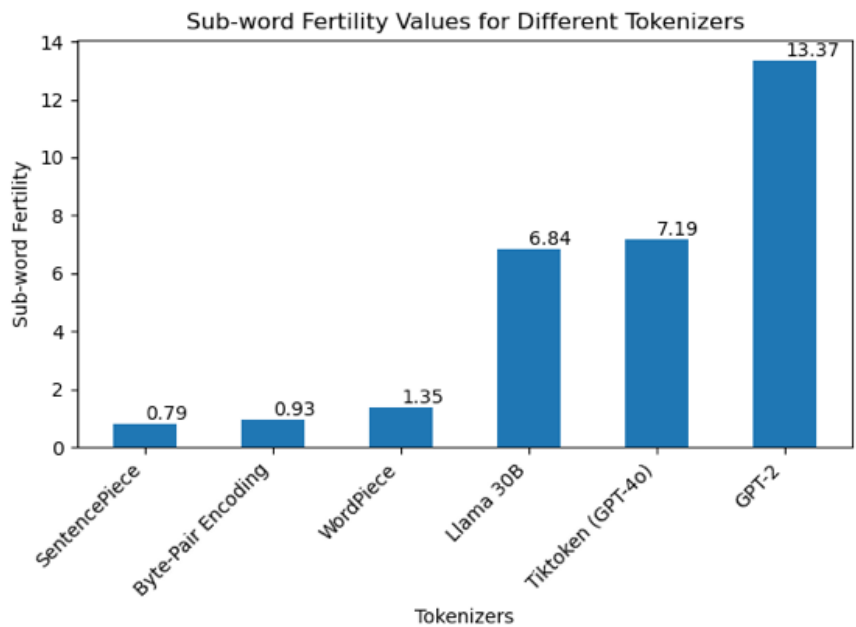

Figure 2. Subword fertility of each tokenizer

The Proportion of Continued Words (PCW) is another useful metric for evaluating the performance of tokenizing algorithms. PCW measures the ratio of words that are split into two or more tokens during tokenization. SentencePiece has the lowest PCW value among the three candidate tokenizers as shown in Table 2. This low PCW value indicates that SentencePiece is more efficient in preserving the integrity of words, as fewer words are broken into multiple tokens as compared to other tokenizers. SentencePiece shows superior also in the tokenization of words.

Table 2 Proportion of continued words

| Tokenizers | Number of continued words | Proportion of continued words |
|---|---|---|
| WordPiece | 49233 | 0.27 |
| Byte-Pair Encoding | 23118 | 0.13 |
| SentencePiece | 15821 | 0.09 |

The total number of words in the dataset is 180000

The last metric used for evaluation is the execution time of each tokenizer. The execution time of baseline tokenizers is also recorded for comparison with the candidate tokenizers. Execution time is the measure of the runtime of the algorithm over each loop. Table 3 and Fig 3 show the execution time for each tokenizer and SentencePiece has the lowest execution time among the tokenizers and indicates that it is the fastest tokenizer. Therefore, SentencePiece also excels in speed making it a suitable candidate for Dzongkha-related NLP tasks.

Table 3 Execution time of each tokenizer encoding the given dataset:

| **Tokenising Algorithms** | **Execution Time** |
|---|---|
| WordPiece | 386 ms ± 6.44 ms per loop (mean ± std. dev. of 7 runs, 1 loop each) |
| SentencePiece (Unigram) | 131 ms ± 2.56 ms per loop (mean ± std. dev. of 7 runs, 10 loops each) |
| Byte-Pair Encoding | 439 ms ± 9.67 ms per loop (mean ± std. dev. of 7 runs, 1 loop each) |
| GPT-2 | 218 ms ± 2.01 ms per loop (mean ± std. dev. of 7 runs, 1 loop each) |
| Tiktoken (GPT-4o) | 205 ms ± 2.82 ms per loop (mean ± std. dev. of 7 runs, 1 loop each) |
| Llama 30B | 690 ms ± 9.67 ms per loop (mean ± std. dev. of 7 runs, 1 loop each) |

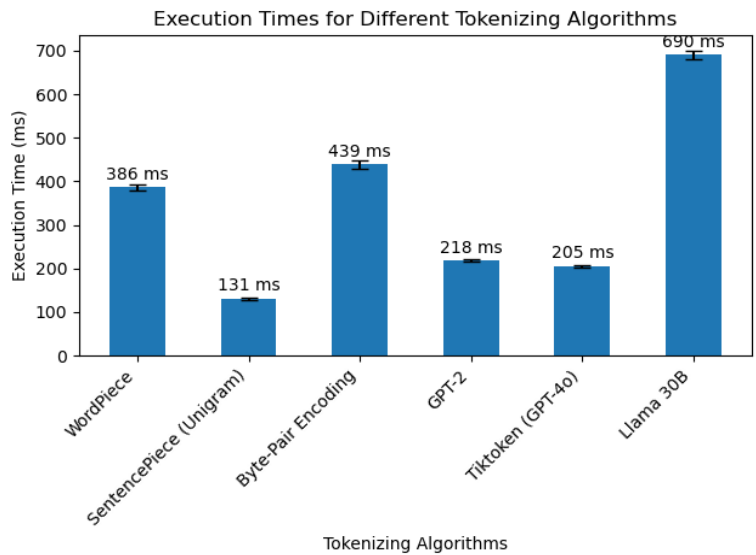

Fig 3. Execution time for each tokenizer

## 6    Discussion

The study shows that SentencePiece is the best-suited tokenizer for Dzongkha. SentencePiece has the best Normalized Sequence Length value, the lowest subword fertility value, excels at the Proportion of Continued Words and has the fastest execution time among the three candidate tokenizers. Although SentencePiece performs worst at the initial pretraining corpus stage, this should not hamper its performance once training is completed. Its superior metrics in other areas make it highly efficient and effective. Based on these results, SentencePiece emerges as the most suitable tokenizer for Dzongkha NLP applications downstream, ensuring both accuracy and efficiency.

Although this study focused on evaluating tokenizers based on specific evaluation metrics, future research should extend evaluation to downstream application tasks such as translation, classification, sentiment analysis and text generation. Such extensions will provide a more comprehensive understanding of the tokenizers' effectiveness and their impact on Dzongkha NLP applications.

## 7    Conclusion

This paper evaluated three popular tokenization algorithms - WordPiece, SentencePiece and Byte-Pair Encoding – for their suitability in downstream tasks related to Dzongkha NLP. Tokenizers are crucial components of NLP and LLMs, but most pretrained tokenizers are not well-suited for low-resource languages like Dzongkha. Developing or training existing tokenizers from scratch is therefore essential.

In this study, we compared the three candidate algorithms against the three baseline tokenizers: Llama 30B, GPT2 and TikToken (GPT-4o). The study indicates that SentencePiece is the most suitable tokenizer for Dzongkha, achieving the best performance in terms of Normalized Sequence Length (NSL), subword fertility, proportion of continued words and execution time.